\documentclass{article}

    \PassOptionsToPackage{numbers, compress}{natbib}

\usepackage[preprint]{neurips_2020}

\usepackage[utf8]{inputenc} 
\usepackage[T1]{fontenc}    
\usepackage{hyperref}       
\usepackage{url}            
\usepackage{booktabs}       
\usepackage{amsfonts}       
\usepackage{nicefrac}       
\usepackage{microtype}      
\usepackage{graphicx}
\usepackage{amsmath}
\usepackage{amsthm}
\usepackage{graphicx}
\usepackage{caption}
\usepackage{subcaption}
\usepackage[final]{pdfpages}

\newcommand{\muzero}{\textit{MuZero} }
\DeclareMathOperator{\E}{\mathbb{E}}

\title{Continuous Control for Searching and Planning \\ with a Learned Model}

\author{%
    Xuxi Yang, Peng Wei \\
    Iowa State University\\
    \texttt{xuxiyang@iastate.edu} \\
    \And
    Werner Duvaud \\
    Ecole Centrale de Nantes \\
    \texttt{werner.duvaud@eleves.ec-nantes.fr} \\
}

\begin{document}

\maketitle

\begin{abstract}
Decision-making agents with planning capabilities have achieved huge success in the challenging domain like Chess, Shogi, and Go. In an effort to generalize the planning ability to the more general tasks where the environment dynamics are not available to the agent, researchers proposed the \muzero algorithm that can learn the dynamical model through the interactions with the environment. In this paper, we provide a way and the necessary theoretical results to extend the \textit{MuZero} algorithm to more generalized environments with continuous action space. Through numerical results on two relatively low-dimensional MuJoCo environments, we show the proposed algorithm outperforms the soft actor-critic (SAC) algorithm, a state-of-the-art model-free deep reinforcement learning algorithm.
\end{abstract}

\section{Introduction}
Planning algorithms based on lookahead search has been very successful for decision-making problems with known dynamics, such as board games \cite{silver2017masteringshogi,silver2016mastering,silver2017mastering} and simulated robot control \cite{tassa2012synthesis}.
However, to apply planning algorithms to more general tasks with unknown dynamics, the agent needs to learn the dynamics model from the interactions with the environment. Although learning the dynamics model has been a long-standing challenging problem, planning with a learned model has several benefits, including data efficiency, better performance, and adaptation to different tasks \cite{hafner2018learning}.
Recently, a model-based reinforcement learning algorithm \textit{MuZero} \cite{schrittwieser2019mastering} was proposed to extend the planning ability to more general environments through learning the dynamics model from the experiences. Building upon \textit{AlphaZero}'s \cite{silver2017mastering} powerful search and search-based policy iteration algorithms, \textit{MuZero} achieves state-of-the-art performance in Atari 2600 with visually rich domains and board games that require precision planning.

However, while \muzero is able to solve problems efficiently with high-dimensional observation spaces, it can only handle environments with discrete action spaces. Many real-world applications, especially physical control problems, require agents to sequentially choose actions from continuous action spaces. While discretizing the action space is a possible way to adapt \textit{MuZero} to continuous control problems, the number of actions increases exponentially with the number of degrees of freedom. Besides, action space discretization can not maintain the information about the structure of the action domain, which may be essential for solving many problems \cite{lillicrap2015continuous}. 

In this paper, we provide a possible way and the necessary theoretical results to extend the \textit{MuZero} algorithm to the continuous action space environments. More specifically, to enable the tree search process to handle the continuous action space, we use \textit{progressive widening} \cite{chaslot2008progressive} strategy which gradually adding actions from the action space to the search tree. 
For the policy parameterization of the policy network output that aims to narrow down the search to high-probability moves, we use the Gaussian distribution to represent the policy and learn statistics of the probability distribution from the experience data \cite{sutton2018reinforcement}. 
For the policy training, a loss function is derived to match the predicted policy output by the policy network and the search policy during the Monte Carlo Tree Search (MCTS) simulation process. Through the above extensions, we show the proposed algorithm in this paper, continuous \textit{MuZero}, outperforms the soft actor-critic method (SAC) \cite{haarnoja2018soft} in relative low-dimensional MuJoCo environments.

This paper is organized as follows. Section 2 presents related work on model-based reinforcement learning and tree search algorithm in continuous action space. In Section 3 we describe the MCTS with the progressive widening strategy in continuous action space. Section 4 covers the network loss function and the algorithm training process. Section 5 presents the numerical experiment results, and Section 6 concludes this paper.

\section{Related Work}
In this section, we briefly review the model-based reinforcement learning algorithms and the Monte Carlo Tree Search (MCTS) in continuous action space.

\subsection{Model-based Reinforcement Learning}
Reinforcement learning algorithms are often divided into model-free and model-based algorithms
\cite{sutton2018reinforcement}. Model-based reinforcement learning algorithms learn a model of the environments, with which they can use for planning and predicting the future steps. This gives the agent an advantage in solving problems requiring a sophisticated lookahead. A classic approach is to directly model the dynamics of the observations \cite{sutton1991dyna,ha2018recurrent,kaiser2019model,hafner2018learning}.
Model-based reinforcement learning algorithm like Dyna-Q \cite{sutton1990integrated,yao2009multi} combines model-free and model-based algorithms using its model to generate samples for model-free algorithms which augment the samples obtained through interaction with the environment. Dyna-Q has been adapted to continuous control problems in \cite{gu2016continuous}. This approach requires learning to reconstruct the observations without distinguishing between useful information and details.

The \muzero algorithm \cite{schrittwieser2019mastering} avoids this by encoding observations into a hidden state without imposing the constraints of capturing all information necessary to reconstruct the original observation, which helps reduce computations by focusing on the information useful for planning. From the encoded hidden state (which has no semantics of environment state attached to it), \muzero also has the particularity of predicting the quantities necessary for a sophisticated lookahead: the policy and value function based on the current hidden state, the reward and next hidden state based on the current hidden state and the selected action. \muzero uses its model to plan with an MCTS search which outputs an improved policy target. It is quite similar to value prediction networks \cite{oh2017value} but uses a predicted policy in addition to the value to prune the search space. However, most of this work is on discrete models but many real-world reinforcement learning domains have continuous action spaces. The most successful methods in domains with a continuous action space remain model-free algorithms \cite{wang2019benchmarking,haarnoja2018soft,schulman2017proximal}.

\subsection{MCTS with Continuous Action Space}
Applying the tree search algorithm to continuous action case causes tension between exploring the larger set of candidate actions to cover more actions, and exploiting the current candidate actions to evaluate them more accurately through deeper search and more execution outcomes. Several recent research works have sought to adapt the tree search to continuous action space.

\cite{tesauro1997line} proposed truncated Monte Carlo that prunes away both candidate actions that are unlikely to be the best action, and the candidates with values close to the current best estimate (i.e., choosing either one would not make a significant difference). Similarly, AlphaGo \cite{silver2016mastering} uses a trained policy network to narrow down the search to high-value actions.
The classical approach of progressive widening (or unpruning) \cite{coulom2007computing,chaslot2008progressive,couetoux2011continuous,yee2016monte} can handle continuous action space by considering a slowly growing discrete set of sampled actions, which has been theoretically analyzed in \cite{wang2009algorithms}.
\cite{mansley2011sample} replaces the Upper Confidence Bound (UCB) method \cite{kocsis2006bandit} in the MCTS algorithm with Hierarchical Optimistic Optimization (HOO) \cite{bubeck2011x}, an algorithm with theoretical guarantees in continuous action spaces. However, the HOO method has quadratic running time, which makes it intractable in games that require extensive planning. In this paper, we apply the progressive widening strategy since for its computation efficiency (it does not increase computation time during the tree search process).

\section{Monte Carlo Tree Search in Continuous Action Space}

We now describe the details of the MCTS algorithm in continuous action space, with a variant of the UCB \cite{kocsis2006bandit} algorithm named PUCB (Predictor + UCB) \cite{rosin2011multi}. UCB has some promising properties: it’s very efficient and guaranteed to be within a constant factor of the best possible bound on the growth of regret (defined as the expected loss due to selecting sub-optimal action), and it can balance exploration and exploitation very well \cite{kocsis2006bandit}. Building upon UCB, PUCB incorporates the prior information for each action to help the agent select the most promising action, which can bring benefits especially for large/continuous action space.

For the MCTS algorithm with a discrete action space, the PUCB score \cite{rosin2011multi} for all actions can be evaluated and the action with the max PUCB score will be selected. However, when the action space becomes large or continuous, it is impossible to enumerate the PUCB score for all possible actions.
Under such scenarios, \textit{progressive widening} strategy deals with the large/continuous action space through artificially limiting the number of actions in the search tree based on the number of visits to the node and slowly growing the discrete set of sampled actions during the simulation process. After the quality of the best available action is estimated well, additional actions are taken into consideration.
More specifically, at the beginning of each action selection step, the algorithm continues by either improving the estimated value of current child actions in the search tree by selecting an action with max PUCB score, or exploring untried actions by adding an action under the current node. This decision is based on keeping the number of child actions for a node bounded by a sublinear function $p({s})$ of the number of visits to the current node denoted as $n({s})$:
\begin{equation}
\label{pw}
    p({s}) = C_{pw} \cdot n({s})^\alpha
\end{equation}
In Eq.~\ref{pw}, $C_{pw}$ and $\alpha \in (0, 1)$ are two parameters that balance whether the MCTS algorithm should cover more actions or improve the estimate of a few actions. At each selection step, if the number of child actions of a node ${s}$ is smaller than $p({s})$, a new action will be added as a child action to the current node. Otherwise, the agent will select an action from the current child actions according to their PUCB score.
PUCB score assures that the tree grows deeper more quickly in the promising parts of the search tree. The progressive widening strategies add that it also grows wider to explore more for some part of the search tree.

To represent the action probability density function in continuous action space, in this paper we let the policy network learn a Gaussian distribution as the policy distribution. The policy network will output ${\mu}$ and ${\sigma}$ as the mean and standard deviation of the normal distribution. With the mean and standard deviation, the action is sampled from the distribution
\begin{equation}
\label{pdf}
    \pi({a} | {s}, \theta) = \frac{1}{\sigma(s, {\theta}) \sqrt{2 \pi}} \exp \left(-\frac{({a} - \mu(s, \theta))^{2}}{2 \sigma(s, \theta)^{2}}\right)
\end{equation}

In the tree search process of the Continuous \textit{MuZero} algorithm, every node of the search tree is associated with a hidden state $h$, either through the pre-processing of the representation network, or through the dynamics network prediction. For each of the action $a$ currently in the search tree (note the number of child actions for each node keeps changing during the simulation process), there is an edge $(s, a)$ that stores a set of statistics ${N(s, a), Q(s, a), \mu(s, a), \sigma(s, a), R(s, a), S(s, a)}$, respectively representing visit counts $N$, mean value $Q$, policy mean $\mu$, policy standard deviation $\sigma$, reward $R$, and state transition $S$. Similar to the \muzero algorithm, the search is divided into three stages, repeated for a number of simulations.

\textbf{Selection:} Each simulation starts from the current root node $s^0$, keeps repeating the selection until reaching an unexpanded node that has no child actions. For each hypothetical time step $k = 1, \cdots$, a decision is made by comparing the number of child actions $|\mathcal{A}_{s^{k-1}}|$ for the node $s^{k-1}$ and the value $p(s^{k-1})$ in Eq.~\ref{pw}.

If $|\mathcal{A}_{s^{k-1}}| \geq p(s^{k-1})$, an action $a^k$ will be selected according to the stored statistics of node $s^{k-1}$, by maximizing over the PUCB score \cite{rosin2011multi,silver2018general}
\begin{equation}
    a^{k}=\arg \max _{a}\left[Q(s, a) + \bar{\pi}(a|s, \theta) \cdot \frac{\sqrt{\sum_{b} N(s, b)}}{1+N(s, a)}\left(c_{1}+\log \left(\frac{\sum_{b} N(s, b)+c_{2}+1}{c_{2}}\right)\right)\right]
\end{equation}
Here we note the difference with the \muzero algorithm is that the prior value here $\bar{\pi}(a|s, \theta)$ is normalized from the policy probability density function value $\pi(a|s, \theta)$ in Eq.~\ref{pdf}, since the density value can be unbounded and PUCB algorithm requires that the prior values are all positive and summed to 1.
\begin{equation}
\label{prior_normalization}
    \bar{\pi}(a|s, \theta) = \frac{{\pi}(a|s, \theta)}{\sum_b {\pi}(b|s, \theta)}
\end{equation}
The constants $c_1$ and $c_2$ are used to control the influence of the prior $\bar{\pi}(a|s, \theta)$ relative to the value $Q(s, a)$, which follow the same parameter setting with \muzero algorithm.


If $|\mathcal{A}_{s^{k-1}}| < p(s^{k-1})$, the agent will select a new action from the action space, add it to the search tree, and expand this new edge. \cite{moerland2018a0c} proposed to sample new a new action according to the mean and standard deviation values output by the policy network and stored at the parent node $s^{k-1}$, which can effectively prune away child actions with low prior value. In this paper, we adopt this naive strategy to focus our work on the first step to extend the \muzero algorithm in continuous action space. We expect the algorithm performance can be further improved with a better action sampling strategy such as \cite{yee2016monte,lee2018deep}.


\textbf{Expansion:} When the agent reaches an unexpanded node either through the PUCB score maximization, or through the progressive widening strategy, the selection stage finishes and the new node will get expanded (we denote this final hypothetical time step during the selection stage as $l$). In the node expansion process, based on the current state-action information $(s^{l-1}, a^l)$ the reward and next state are first computed by the dynamics function, $r^{l}, s^{l} = g_{\theta}(s^{l-1}, a^{l})$. With the state information $s^l$ for the next step, the policy and value are then computed by the prediction function, ${\mu}^{l}, {\sigma}^l, v^{l}=f_{\theta}(s^{l})$. 
After the function prediction, the new node corresponding to state $s^{l}$ is added to the search tree under the edge $(s^{l-1}, a^l)$.

In the \textit{MuZero} algorithm, with a finite number of actions, each edge $(s^{l}, a)$ from the newly expanded node is initialized according to the predicted policy distribution.
Since the action space is now continuous, only one edge with the action value randomly sampled from $\mathcal{N}(\mu^l, (\sigma^l)^2)$ is added to the newly expanded node, and the statistics for this edge is initilaized to ${N({s}, {a}) = 0, Q({s}, {a}) = 0, \mu({s}, {a}) = {\mu}, \sigma({s}, {a}) = {\sigma}}$, where $\mu$ and $\sigma$ are used to determine the probability density prior for the sampled $a$.
Similar to the \muzero and \textit{AlphaZero} algorithm, the search algorithm with progressive strategy makes at most one call to the dynamics function and prediction function respectively per simulation, maintaining the same order of computation cost.

\textbf{Backup:} In the general MCTS algorithm, there is a simulation step that performs one random playout from the newly expanded node to a terminal state of the game. However, in the \muzero algorithm, each step from an unvisited node will require one call to the dynamics function and the prediction function, which makes it intractable for games that need a long trajectory to finish. Thus similar to the \muzero algorithm, immediately after the Expansion step, the statistics (the mean value $Q$ and the visit count $N$) for each edge in the simulation path will be updated based on the cumulative discounted reward, bootstrapping from the value function of the newly expanded node.


\section{Neural network training in continuous action space}

In the \muzero algorithm, the parameters of the representation, dynamics, and prediction networks are trained jointly, through backpropagation-through-time, to predict the policy, the value function, and the reward. At each training step, a trajectory with $K$ consecutive steps are sampled from the replay buffer, from which the network targets for the policy, value, and reward are calculated. The loss functions for the policy, value, and reward are designed respectively to minimize the difference between the network predictions and targets. In the following, we describe how we design the loss function for the policy network in continuous action space, and briefly review the loss function for the value/reward network.

\subsection{Policy Network}

Similar to other policy-based algorithms \cite{schulman2015trust,duan2016benchmarking}, the policy network outputs the mean $\mu(s, \theta)$ and the standard deviation $\sigma(s, \theta)$ of a Gaussian distribution. For the policy mean, we use a fully-connected MLP with leakyrelu nonlinearities for the hidden layers and tanh activation for the output layer. A separate fully-connected MLP specifies the log standard deviation, which also depends on the state input.


For the training target of the policy network, we want to transform the MCTS result of the root node to a continuous target probability density $\hat{\pi}$.
In general, to estimate the density of a probability distribution with continuous support, independent and identically distributed (i.i.d.) samples are drawn from the underlying distribution \cite{perez2008kullback}. 
However, in the MCTS algorithm, only a finite number of actions with different visit counts will be returned (for MCTS algorithm with $N$ simulations, the number of actions will be $\left \lceil{C_{pw} \cdot N^{\alpha}}\right \rceil$). Thus here we assume the density value of the target distribution $\hat{\pi}(a | {s})$ at a root action ${a}_i$ is proportional to its visit counts
\begin{equation}
\label{density_norm}
    \hat{\pi}(a_{i} | {s})=\frac{n({s}, {a}_i)^{\tau}}{Z({s}, \tau)}
\end{equation}
where $\tau \in \mathbb{R}^{+}$ specifies the temperature parameter, and $Z({s}, \tau)$ is a normalization term that only depends on the current state and the temperature parameter. 

\begin{figure}
  \centering
  \includegraphics[width=.7\textwidth]{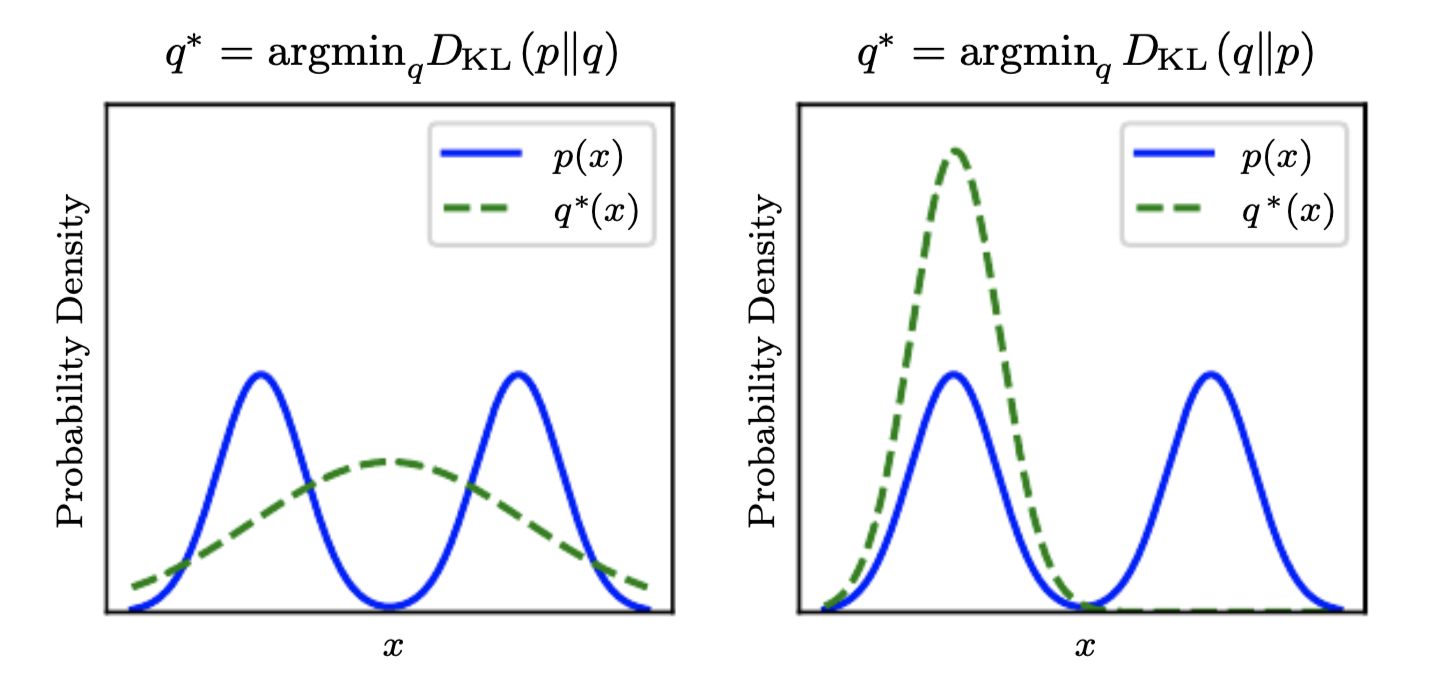}
  \caption{Illustration of the KL divergence between the prediction and target.}
  \label{kl}
\end{figure}

For the loss function, we use the Kullback-Leibler divergence between the network output $\pi_{\theta}(a | s)$ and the empirical density $\hat{\pi}({a} | {s})$ from the MCTS result.
\begin{equation}
\label{kl_loss}
  l^p (\theta) = \mathrm{D}_{KL}(\pi_{\theta}({a}| {s}) \| \hat{\pi}({a} | {s}))
\end{equation}
In general, since the KL divergence between two distributions is asymmetric, we have the choice of minimizing either $\mathrm{D}_{KL}(\pi_{\theta} \| \hat{\pi})$ or $\mathrm{D}_{KL}(\hat{\pi} \| \pi_{\theta})$. As illustrated in Fig.~\ref{kl} from \cite{Goodfellow-et-al-2016}, the choice of the direction is problem dependent. The loss function $\mathrm{D}_{KL}(\pi_{\theta} \| \hat{\pi})$ require $\pi_\theta$ to place high probability anywhere that the $\hat{\pi}$ places high probability, while the other loss function requires $\pi_\theta$ to rarely places high probability anywhere that $\hat{\pi}$ places low probability. As in our problem, we desire the trained policy network to prune the undesired actions with low return, thus we choose $\mathrm{D}_{KL}(\pi_{\theta} \| \hat{\pi})$ as the policy loss function.

Also note that the empirical density $\hat{\pi}({a} | {s})$ from the search result does not define a proper density, as we never specify the density value in between the finite support points. However, through the following theorem, we show that even if we only consider the loss at the support points as show in Eq.~\ref{hat_policy_loss}, the expectation of loss function equals the true KL divergence if we sample actions according to the policy network prediction $a \sim \pi_{\theta}(a|s)$.
\begin{equation}
\label{hat_policy_loss}
    \hat{l}_N^p(\theta) = \frac{1}{N} \sum_i^N
    \log \pi_{\theta}(a_i | {s}) - \log \hat{\pi}_{\theta}(a_i | {s})
\end{equation}

\newtheorem{th1}{Theorem}
\begin{th1}
If the actions are sampled according to $\pi_\theta(a | {s})$, then for the empirical estimator $\hat{l}_N^p(\theta)$ of the policy loss function, we have
\begin{equation}
\label{th1}
  \E_{a \sim \pi_\theta(a | {s})} \left[\hat{l}_N^p(\theta) \right] = \mathrm{D}_{KL}(\pi_{\theta}(a| {s}) \| \hat{\pi}(a | {s}))
\end{equation}
Further more, the variance of the empirical estimator $\hat{l}_N^p(\theta)$ converges to 0 with $O(N)$.
\end{th1}

We provide the proof in the Appendix. Theorem 1 states that if $a \sim \pi_\theta(a | {s})$, then the empirical estimator $\hat{l}_N^p(\theta)$ for the policy loss function in Eq.~\ref{kl_loss} is an unbiased estimator.

By subtituting Eq.~\ref{density_norm} into the estimator in Eq.~\ref{hat_policy_loss}, we can further simplify the estimator:
\begin{equation}
\begin{split}
  \hat{l}_N^p(\theta) 
  &= \frac{1}{N} \sum_i^N \left( \log \pi_{\theta}(a_i | {s}) - \log \hat{\pi}_{\theta}(a_i | {s}) \right) \\
  &= \frac{1}{N} \sum_i^N \left( \log \pi_{\theta}(a_i | {s}) - \log \frac{n({s}, {a}_{{i}})^{\tau}}{Z({s}, \tau)} \right)    \\
  &= \frac{1}{N} \sum_i^N \left( \log \pi_{\theta}(a_i | {s}) - \log n({s}, {a}_{{i}})^{\tau} \right) + \log Z({s}, \tau)
\end{split}
\end{equation}
where the term $\log Z({s},\tau)$ can be dropped since it does not depend on neural network weights $\theta$ and action $a$, which means it is a constant given a specific data sample. Thus the policy loss function becomes
\begin{equation}
\label{final_policy_loss}
  \tilde{l}_N^p(\theta) = \frac{1}{N} \sum_i^N \left( \log \pi_{\theta}(a_i | {s}) - \log n({s}, {a}_{{i}})^{\tau} \right)
\end{equation}

We also give the derivation of the expected gradient for the empirical loss $\tilde{l}_N^p(\theta)$, with the details provided in the Appendix:
\begin{equation}
  \E_{a \sim \pi_{\theta}(a | {s})} \nabla_{\theta} \tilde{l}_N^p(\theta) = \E_{a \sim \pi_{\theta}(a | {s})} \left[\nabla_{\theta} \log \pi_\theta(a| {s}) \cdot (\log \pi_\theta(a| {s}) - \tau \log n(a, {s})) \right]
\end{equation}

Also note here that for the estimator $\hat{l}_N^p(\theta)$ to be unbiased, the actions need to be sampled according to the distribution $\pi_\theta$, which depends on $\mu_\theta$ and $\sigma_\theta$ predicted from the policy network. However, for the experience data sampled from the replay buffer, the actions are sampled according to old policy network prediction during the self play phase. In this paper, we replace the expectation over ${a} \sim \pi_{\theta}({a} | {s})$ with the empirical support points from old policy distribution ${a} \sim \pi_{\theta_{old}}$, where $\theta_{old}$ denotes the old network weights during the self play phase. Although the empirical estimate become biased with this replacement, it does not affect the performance of the proposed algorithm from the numerical experiments results. Future work beyond this paper would include using weighted sampling to determine an unbiased estimation of the policy loss. The final policy loss function and its gradient become
\begin{equation}
  \hat{l}^p(\theta) = \E_{{a} \sim \pi_{\theta_{old}}} \left( \log \pi_{\theta}(a_i | {s}) - \tau \log n({s}, {a}_{{i}}) \right)
\end{equation}
\begin{equation}
  \nabla_{\theta} \hat{l}^p(\theta) =\E_{{a} \sim \pi_{\theta_{old}}}
  [\nabla_{\theta} \log \pi_\theta({a}| {s}) \cdot (\log \pi_\theta({a}| {s}) - \tau \log n({a}, {s}))]
\end{equation}



\subsection{Value/Reward Network}
For the value/rework network, we follow the same setting with the \muzero algorithm, and we briefly review the details in this subsection.
Following \cite{pohlen2018observe}, the value/reward targets are scaled using an invertible transform 
\begin{equation}
\label{transform}
  h(x) = \operatorname{sign}(x)(\sqrt{|x|+1}-1)+\epsilon x
\end{equation}
where $\epsilon=0.001$ in our experiments. We then apply a transformation $\phi$ to the scalar reward and value targets to obtain equivalent categorical representations. For the reward/value target, we use a discrete support set of size 21 with one support for every integer between -10 and 10. Under this transformation, each scalar is represented as the linear combination of its two adjacent supports, such that the original value can be recovered by $x = x_{\text{low}} \cdot p_{\text{low}} + x_{\text{high}} \cdot p_{\text{high}}$. 

During inference the actual value and rewards are obtained by first computing their expected value under their respective softmax distribution and subsequently by inverting the scaling transformation using Eq.~\ref{transform}. Scaling and transformation of the value and reward happen transparently on the network side and is not visible to the rest of the algorithm.

With the above formulation, the loss for the value/reward has the following form
\begin{equation}
    l^{v}(z, \mathbf{q}) = \boldsymbol{\phi}(z)^{T} \log \mathbf{q}
\end{equation}
\begin{equation}
    l^r(u, \mathbf{r}) = \boldsymbol{\phi}(u)^{T} \log \mathbf{r}
\end{equation}
where $z, u$ are the value/reward targets, $\mathbf{q}, \mathbf{r}$ are the value/reward network output, and $\boldsymbol{\phi}$ denotes the transformation from the scalar values to the categorical representations.

\subsection{Loss Function}
For reinforcement learning algorithms in continuous action space, there is a risk that the policy network may converge prematurely, hence losing any exploration \cite{haarnoja2018soft}. To learn a policy that acts as randomly as possible while still being able to succeed at the task, in the numerical experiment we also add an entropy loss to the policy:
\begin{equation}
    l^h(\theta) = -H(\pi_{\phi}(a | s))
\end{equation}

With the above formulation, the loss function for the proposed continuous \textit{MuZero} algorithm is a weighted sum of the loss functions described above, with a weight normalization term:
\begin{equation}
    l(\theta) =
    l^{r} +
    l^{v} +
    \tilde{l}_N^p +
    \lambda \cdot l^{h} +
    c \cdot \|\theta\|^{2}
\end{equation}
where $\lambda$ controls the contribution of the entropy loss and $c$ is the coefficient for the weight normalization term.

\subsection{Network Training}
The original version of the \muzero algorithm uses a residual network which is suitable for image processing in Atari environments and board games. In our experiments with MuJoCo environments, we replace the networks (policy network, value network, reward network, and dynamics network) with fully connected networks. The hyperparameter and network architecture details are provided in the Appendix.

During the data generation process, we use 3 actors deployed on the CPU to keep generating experience data using the proposed algorithm, by pulling the most recent network weights from time to time. The same exploration scheme with the \muzero algorithm is used, where the visit count distribution is parametrized using a temperature parameter $T$ that can balance the exploitation and exploration.

At each training step, an episode is first sampled from the replay buffer, and then $K$ consecutive transitions are sampled from the episode. During the sampling process, the samples are drawn according to prioritized replay \cite{schaul2015prioritized,brittain2019prioritized}. The priority for transition $i$ is $P(i)=\frac{p_{i}^{\alpha}}{\sum_{k} p_{k}^{\alpha}}$, where $p_i$ is determined through the difference between the search value and the observed n-step return. The priority for an episode equals the mean priorities of all the transitions in this episode. To correct for sampling bias introduced by the prioritized sampling, we scale the loss using the importance sampling ratio $w_{i}=\left(\frac{1}{N} \cdot \frac{1}{P(i)}\right)^{\beta}$.

To maintain a similar magnitude of the gradient across different unroll steps, we scale the gradient following the \muzero algorithm.
To improve the learning process and bound the activation function output, we scale the hidden state to the same range as
the action input ($[-1, 1]$):
\begin{equation}
  s_{\text {scaled}}= \frac{2s- [\min (s) + \max(s)] }{\max (s)-\min (s)}
\end{equation}

\section{Experimental Results}
In this section, we show the preliminary experimental results on two relatively low-dimensional MuJoCo environments, compared with Soft Actor-Critic (SAC), a state-of-the-art model-free deep reinforcement learning algorithm. For the comparison with the SAC algorithm, the stable baselines \cite{stable-baselines} implementation was used, with the same parameter setting following the SAC paper \cite{haarnoja2018soft}.

\begin{figure}
\centering
\begin{subfigure}{.5\textwidth}
  \centering
  \includegraphics[width=.99\linewidth]{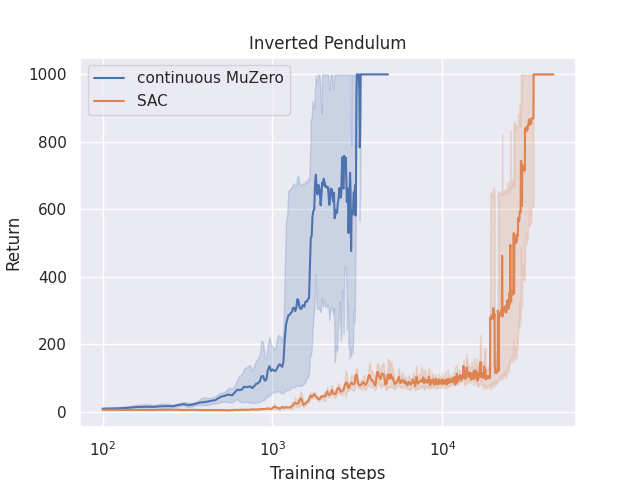}
  \caption{Converged after 4k training steps.}
  \label{fig:sub1}
\end{subfigure}%
\begin{subfigure}{.5\textwidth}
  \centering
  \includegraphics[width=.99\linewidth]{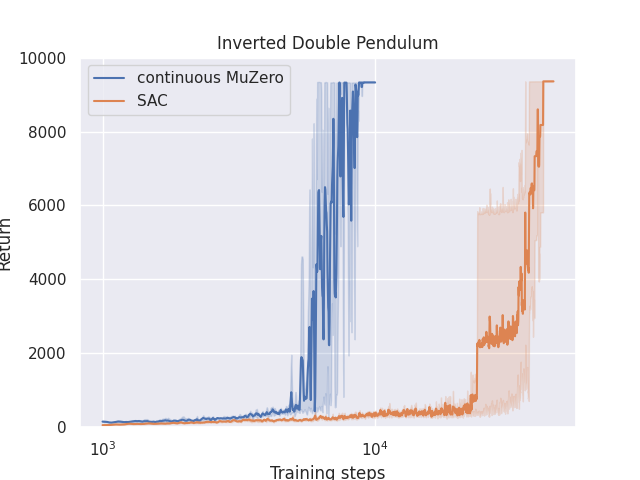}
  \caption{Converged after 9k training steps.}
  \label{fig:sub2}
\end{subfigure}
\caption{Episode rewards achieved during the training process averaged over 5 random seeds. Our proposed algorithm (blue line) achieves better score at earlier training steps than SAC (orange line). Note the log scale on the x-axis.}
\label{IDP}
\end{figure}

\begin{figure}
\centering
\begin{subfigure}{.5\textwidth}
  \centering
  \includegraphics[width=.99\linewidth]{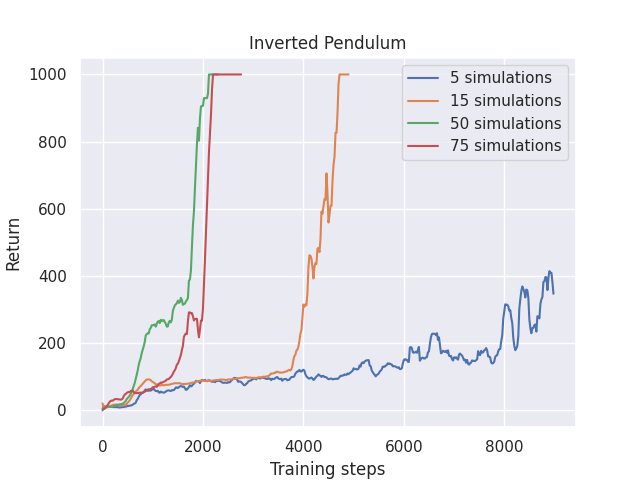}
  \label{fig:sub3}
\end{subfigure}%
\begin{subfigure}{.5\textwidth}
  \centering
  \includegraphics[width=.99\linewidth]{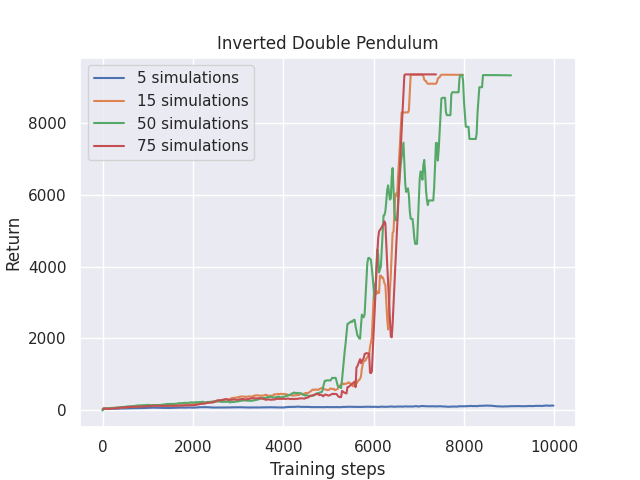}
  \label{fig:sub4}
\end{subfigure}
\caption{Performance of the proposed algorithm with different number of simulations, with more simulations resulting in better performance.}
\label{simulations-results}
\end{figure}

We conducted experiments on InvertedPendulum-v2 and InvertedDoublePendulum-v2 tasks over 5 random seeds, and the results are shown in Fig.~\ref{IDP}. From this plot we can see the proposed continuous \textit{MuZero} algorithm consistently outperforms the SAC algorithm. Our proposed algorithm converges to the optimal score after training for 4k steps for Inverted Pendulum and 9k steps for Inverted Double Pendulum, achieves better data efficiency.

In Fig.~\ref{simulations-results}, we also varied the number of simulations in the experiments to illustrate its effect on the continuous \muzero algorithm. The algorithm trained and played with more simulations is able to converge faster, which corresponds to our intuition, since
the simulation number determines the size of the search tree, where a higher number allows the action space to be explored more (resulting in a wider tree) and estimated with greater precision (resulting in a deeper tree), at the cost of more intensive calculations.


\section{Conclusion}
This paper provides a possible way and the necessary related theoretical results to extend the \muzero algorithm to continuous action space environments. We propose a loss function for the policy in continuous action case that can help the policy network to match the search results of the MCTS algorithm. The progressive widening is used to gradually extend the action space, which is an effective strategy to deal with large/continuous action space. Preliminary results on low-dimensional MuJoCo environments show that our approach performs much better than the soft actor-critic (SAC) algorithm. Future work will further explore the empirical performance of the continuous \muzero algorithm on MuJoCo environments with higher dimensions, since the adaption to the \muzero algorithm proposed in this paper can be easily extended to higher dimension action space. Improving the selection process in the MCTS progressive widening process could also be a future direction to help speed up the algorithm convergence.

\section*{Broader Impact}

In this paper, we introduce continuous \muzero algorithm, that achieves the state-of-the-art across some low dimensional continuous control tasks.
Although the experiments are for MuJoCo tasks, we broaden our focus to consider the longer-term impacts of developing decision-making agents with planning capabilities. Such capabilities could be applied to a range of domains, such as robotics, games, business management, finance, and transportation, etc.
Improvements to decision-making strategy likely have complex effects on welfare, depending on how these capabilities are distributed and the character of the strategic setting. For example, depending on who can use this scientific advance, such as criminals or well-motivated citizens, this technology may be socially harmful or beneficial. 

\bibliography{main}
\bibliographystyle{plain}



\section*{Supplementary material}

# Appendix

## 1 Proof of Theorem 1

**Theorem 1.** *If the actions are sampled according to $\pi_\theta(a|s)$, then for the empirical estimator $\hat{l}_N^p(\theta)$ of the policy loss function*

$$\hat{l}_N^p(\theta) = \frac{1}{N}\sum_i^N \log \pi_\theta(a_i|s) - \log \hat{\pi}_\theta(a_i|s) \tag{1}$$

*we have*

$$\mathbb{E}_{a \sim \pi_\theta(a|s)}\left[\hat{l}_N^p(\theta)\right] = \mathrm{D}_{KL}(\pi_\theta(a|s)\|\hat{\pi}(a|s)) \tag{2}$$

*Further more, the variance of the empirical estimator $\hat{l}_N^p(\theta)$ converges to 0 with $O(N)$.*

*Proof.* In the above theorem, the actions are in one-dimensional space. However, the proof is also valid for actions in higher-dimensional space. Thus in the following proof, we will use $\boldsymbol{a}$ in vector form to provide a more general proof.

We simplify the left hand side of Eq. 2.

$$\begin{aligned}
\mathbb{E}\left[\hat{l}_N^p(\theta)\right] &= \frac{1}{N}\mathbb{E}\left[\sum_i^N \left(\log \pi_\theta(\boldsymbol{a}_i|\boldsymbol{s}) - \log \hat{\pi}_\theta(\boldsymbol{a}_i|\boldsymbol{s})\right)\right] \\
&= \frac{1}{N}\sum_i^N \mathbb{E}\left[\frac{\log \pi_\theta(\boldsymbol{a}_i|\boldsymbol{s})}{\log \hat{\pi}_\theta(\boldsymbol{a}_i|\boldsymbol{s})}\right]
\end{aligned} \tag{3}$$

Since we assume that $\boldsymbol{a}_i \sim \pi_\theta(\boldsymbol{a}|\boldsymbol{s})$, we have

$$\begin{aligned}
&\frac{1}{N}\sum_i^N \mathbb{E}\left[\frac{\log \pi_\theta(\boldsymbol{a}_i|\boldsymbol{s})}{\log \hat{\pi}_\theta(\boldsymbol{a}_i|\boldsymbol{s})}\right] \\
=&\frac{1}{N}\sum_i^N \int_{\boldsymbol{a}} \pi_\theta(\boldsymbol{a}|\boldsymbol{s})\frac{\log \pi_\theta(\boldsymbol{a}|\boldsymbol{s})}{\log \hat{\pi}_\theta(\boldsymbol{a}|\boldsymbol{s})}d\boldsymbol{a} \\
=&\int_{\boldsymbol{a}} \pi_\theta(\boldsymbol{a}|\boldsymbol{s})\frac{\log \pi_\theta(\boldsymbol{a}|\boldsymbol{s})}{\log \hat{\pi}_\theta(\boldsymbol{a}|\boldsymbol{s})}d\boldsymbol{a} \\
=&\mathrm{D}_{KL}(\pi_\theta(\boldsymbol{a}|\boldsymbol{s})\|\hat{\pi}(\boldsymbol{a}|\boldsymbol{s}))
\end{aligned} \tag{4}$$

from which we show the unbiasedness of the estimator $\hat{l}_N^p(\theta)$.

Furthermore, due to the Strong Law of Large Numbers, as we increase the number of samples $N$, the estimator $\hat{l}_N^p(\theta)$ becomes a closer and closer approximation:

$$\Pr\left\{\lim_{N \to \infty} \hat{l}_N^p(\theta) = \mathrm{D}_{KL}(\pi_\theta(\boldsymbol{a}|\boldsymbol{s})\|\hat{\pi}(\boldsymbol{a}|\boldsymbol{s}))\right\} = 1 \tag{5}$$

For the variance of the estimator, we have

$$\begin{aligned}
\mathrm{Var}\left(\hat{l}_N^p(\theta)\right) &= \mathrm{Var}\left(\frac{1}{N}\sum_i^N \frac{\log \pi_\theta(\boldsymbol{a}_i|\boldsymbol{s})}{\log \hat{\pi}_\theta(\boldsymbol{a}_i|\boldsymbol{s})}\right) \\
&= \frac{1}{N^2}\sum_i^N \mathrm{Var}\left(\frac{\log \pi_\theta(\boldsymbol{a}_i|\boldsymbol{s})}{\log \hat{\pi}_\theta(\boldsymbol{a}_i|\boldsymbol{s})}\right) \\
&= \frac{1}{N}\mathrm{Var}\left(\frac{\log \pi_\theta(\boldsymbol{a}|\boldsymbol{s})}{\log \hat{\pi}_\theta(\boldsymbol{a}|\boldsymbol{s})}\right)
\end{aligned} \tag{6}$$

For a given state $\boldsymbol{s}$, $\mathrm{Var}\left(\frac{\log \pi_\theta(\boldsymbol{a}|\boldsymbol{s})}{\log \hat{\pi}_\theta(\boldsymbol{a}|\boldsymbol{s})}\right)$ is fixed and we denote it as $\mathrm{Var}(Y)$, thus we have

$$\mathrm{Var}\left(\hat{l}_N^p(\theta)\right) = \frac{1}{N}\mathrm{Var}(Y) \tag{7}$$

which converges to 0 as $N \to \infty$. □

## 2 Derivations of the policy loss gradient

Here we give the derivation of the expected gradient for the empirical loss $\tilde{l}_N^p(\theta)$:

$$\mathbb{E}_{a \sim \pi_\theta(a|s)} \nabla_\theta \tilde{l}_N^p(\theta) = \mathbb{E}_{a \sim \pi_\theta(a|s)} \left[ \nabla_\theta \log \pi_\theta(a|s) \cdot (\log \pi_\theta(a|s) - \tau \log n(s, a) \right] \tag{8}$$

Similar to the proof of Theorem 1, this derivation is also valid for actions in higher dimensional space, and we will use $s$ and $a$ in vector form to provide a more general derivation.

From

$$\tilde{l}_N^p(\theta) = \frac{1}{N} \sum_i^N \left( \log \pi_\theta(\boldsymbol{a}_i|\boldsymbol{s}) - \log n(\boldsymbol{s}, \boldsymbol{a_i})^\tau \right) \tag{9}$$

we have

$$\begin{aligned}
\mathbb{E}_{\boldsymbol{a} \sim \pi_\theta(\boldsymbol{a}|\boldsymbol{s})} \nabla_\theta \tilde{l}_N^p(\theta) &= \mathbb{E}_{\boldsymbol{a} \sim \pi_\theta(\boldsymbol{a}|\boldsymbol{s})} \nabla_\theta \frac{1}{N} \sum_{i=1}^N [\log \pi_\theta(\boldsymbol{a}_i|\boldsymbol{s}) - \tau \log n(\boldsymbol{s}, \boldsymbol{a}_i)] \\
&= \frac{1}{N} \sum_{i=1}^N \nabla_\theta \int_{\boldsymbol{a}} \pi_\theta(\boldsymbol{a}|\boldsymbol{s}) [\log \pi_\theta(\boldsymbol{a}|\boldsymbol{s}) - \tau \log n(\boldsymbol{s}, \boldsymbol{a}_i)] d\boldsymbol{a} \\
&= \int_{\boldsymbol{a}} \log \pi_\theta(\boldsymbol{a}|\boldsymbol{s}) \cdot \nabla_\theta \pi_\theta(\boldsymbol{a}|\boldsymbol{s}) + \pi_\theta(\boldsymbol{a}|\boldsymbol{s}) \cdot \nabla_\theta \log \pi_\theta(\boldsymbol{a}|\boldsymbol{s}) \\
&\quad - \nabla_\theta \pi_\theta(\boldsymbol{a}|\boldsymbol{s}) \cdot \tau \log n(\boldsymbol{s}, \boldsymbol{a}_i) d\boldsymbol{a} \\
&= \int_{\boldsymbol{a}} \pi_\theta(\boldsymbol{a}|\boldsymbol{s}) \cdot \nabla_\theta \log \pi_\theta(\boldsymbol{a}|\boldsymbol{s}) \cdot (\log \pi_\theta(\boldsymbol{a}|\boldsymbol{s}) - \tau \log n(\boldsymbol{s}, \boldsymbol{a}_i)) d\boldsymbol{a} \\
&= \mathbb{E}_{\boldsymbol{a} \sim \pi_\theta(\boldsymbol{a}|\boldsymbol{s})} [\nabla_\theta \log \pi_\theta(\boldsymbol{a}|\boldsymbol{s}) \cdot (\log \pi_\theta(\boldsymbol{a}|\boldsymbol{s}) - \tau \log n(\boldsymbol{s}, \boldsymbol{a}_i))]
\end{aligned} \tag{10}$$

which follows directly from the fact that

$$\nabla_\theta \pi_\theta(\boldsymbol{a}|\boldsymbol{s}) = \pi_\theta(\boldsymbol{a}|\boldsymbol{s}) \cdot \nabla_\theta \log \pi_\theta(\boldsymbol{a}|\boldsymbol{s}) \tag{11}$$

and

$$\begin{aligned}
&\int_{\boldsymbol{a}} \pi_\theta(\boldsymbol{a}|\boldsymbol{s}) \nabla_\theta \log \pi_\theta(\boldsymbol{a}|\boldsymbol{s}) d\boldsymbol{a} \\
&= \int_{\boldsymbol{a}} \pi_\theta(\boldsymbol{a}|\boldsymbol{s}) \cdot \frac{\nabla_\theta \pi_\theta(\boldsymbol{a}|\boldsymbol{s})}{\pi_\theta(\boldsymbol{a}|\boldsymbol{s})} d\boldsymbol{a} \\
&= \int_{\boldsymbol{a}} \nabla_\theta \pi_\theta(\boldsymbol{a}|\boldsymbol{s}) d\boldsymbol{a} \\
&= \nabla_\theta \int_{\boldsymbol{a}} \pi_\theta(\boldsymbol{a}|\boldsymbol{s}) d\boldsymbol{a} \\
&= \nabla_\theta 1 = 0
\end{aligned} \tag{12}$$

## 3 Experiment parameters

Here we provide the details of the hyperparameters we used in the experiments. For both two environments `InvertedPendulum-v2` and `InvertedDoublePendulum-v2`, we use the same set of parameters. For the progressive widening parameter $\alpha$, we set it to 0.49. In this case, when the agent does 50 simulations at each decision-making step, in total 7 actions are sampled from the root state, and each child action, as a root of the subtree, has roughly 7 simulations. To speed up the convergence, we scaled the reward by 1/20 similar to the soft actor-critic algorithm. The smaller value/reward scale enables us to use a smaller value/reward support size of 21 compared to the *MuZero* algorithm. For the network, we used relatively simple network architecture with single 64 neurons hidden layer for the dynamics/reward/value/policy mean network, and we did not include hidden layers for the representation/policy log std network. For the optimizer, we observed that the `Adam` optimizer converges faster compared to the Stochastic Gradient Descent algorithm. The learning rate is kept fixed to be 0.003 during the whole training process. Additionally, to avoid network overfitting, we keep the ratio that the number of self-played games/training steps fixed to 1/5.

| Hyperparameter | Value |
| --- | --- |
| Number of simulations | 50 |
| Discount factor $\gamma$ | 0.997 |
| PUCB $c_1$ | 1.25 |
| PUCB $c_2$ | 19652 |
| Progressive widening $\alpha$ | 0.49 |
| Progressive widening $C_{pw}$ | 1 |
| Support size | 21 |
| Encoding size | 15 |
| Batch size | 128 |
| Entropy loss weight $\lambda$ | 0.1 |
| Optimizer | Adam |
| Weight decay | $10^{-4}$ |
| Learning rate | 0.003 |
| Representation network hidden layer size | None |
| Dynamics network hidden layer size | [64] |
| Reward network hidden layer size | [64] |
| Value network hidden layer size | [64] |
| Policy mean network hidden layer size | [64] |
| Policy log std network hidden layer size | None |
| Replay buffer size | 700 |
| Number of unroll steps $K$ | 10 |
| Number of TD steps | 50 |
| Prioritized replay $\alpha$ | 0.1 |
| Prioritized replay $\beta$ | 1 |
| Self played games per training steps | 1/5 |
| Reward scale | 1/20 |
| Temperature $T$ | 1 |



\end{document}